\documentclass[runningheads]{llncs}
\usepackage{graphicx}

\usepackage{amsmath}
\usepackage{algorithm}
\usepackage{algorithmic}
\usepackage{multirow}
\usepackage{subfig}
\usepackage{float}
\newcommand{\mc}{\mathcal}

\begin{document}
\title{Confidence-Based Task Prediction in Continual Disease Classification Using Probability Distribution}
\titlerunning{Confidence-Based Task Prediction}
%
\author{Tanvi Verma\inst{1} \and
Lukas Schwemer\inst{2} \and
Mingrui Tan\inst{1} \and
Fei Gao\inst{1} \and \\
Yong Liu\inst{1} \and
Huazhu Fu\inst{1} }

\authorrunning{Verma et al.}
%
\institute{Institute of High Performance Computing (IHPC), Agency for Science, Technology and Research (A*STAR), Singapore \and
Karlsruhe Institute of Technology, Germany
\\
\email{vermat@ihpc.a-star.edu.sg}}

\maketitle              
\begin{abstract}
Deep learning models are widely recognized for their effectiveness in identifying medical image findings in disease classification. However, their limitations become apparent in the dynamic and ever-changing clinical environment, characterized by the continuous influx of newly annotated medical data from diverse sources. In this context, the need for continual learning becomes particularly paramount, not only to adapt to evolving medical scenarios but also to ensure the privacy of healthcare data. In our research, we emphasize the utilization of a network comprising expert classifiers, where a new expert classifier is added each time a new task is introduced. We present CTP, a task-id predictor that utilizes confidence scores, leveraging the probability distribution (logits) of the classifier to accurately determine the task-id at inference time. Logits are adjusted to ensure that classifiers yield a high-entropy distribution for data associated with tasks other than their own. By defining a noise region in the distribution and computing confidence scores, CTP achieves superior performance when compared to other relevant continual learning methods. Additionally, the performance of CTP can be further improved by providing it with a continuum of data at the time of inference.
\end{abstract}
\section{Introduction}

Continual learning (CL) methods for disease classification allow deep learning models to adapt to new data and improve their performance over time while trying to preserve previously learned knowledge. These methods are particularly important in medical domain, as the data is often highly dynamic and new information is constantly being discovered. 
Task incremental learning (TIL) and class incremental learning (CIL) are two main scenarios of CL \cite{van2019three}, with TIL involving the availability of task-ids during inference, while CIL lacks such information, making it a more challenging problem to address.
The primary challenge in CL is \textit{catastrophic forgetting} \cite{mccloskey1989catastrophic}, which occurs when the learning of a new task substantially alters the network weights acquired for previous tasks, leading to a decline in accuracy for those old tasks. Furthermore, due to the inability of existing CL methods to achieve comparable accuracy to retraining with the entire dataset (referred as joint training), the technology is not yet fully prepared for medical applications \cite{derakhshani2022lifelonger,verma2023continual,kumari2023continual,jin2023privacy}. One approach to address catastrophic forgetting is to train separate expert models for each task and utilize the appropriate model at inference time. However, determining the correct model to use at inference becomes challenging. Therefore, in this study, we concentrate on the task identification problem in continual learning, aiming to determine the task-id at the time of inference from a network of expert classifiers, a principle also supported in \cite{hinton2015distilling,rusu2016progressive,aljundi2017expert}. The rationale behind this is that when the task identity is either known or can be correctly inferred, employing task-specific expert models, rather than trying to fine-tune a single model for every task, can lead to improved accuracy.


To address the issue of catastrophic forgetting in continual learning, several methods involve storing a small amount of old data, known as exemplars, from previous tasks. These exemplars are then utilized to fine-tune the model on the current task, helping to mitigate interference with previously learned tasks \cite{rebuffi2017icarl,lopez2017gradient,guo2020improved}. However, it's worth noting that storing exemplars may not always be feasible, particularly due to privacy concerns in medical domain. 
To tackle these challenges, we introduce CTP (Confidence-based Task-id Prediction), a novel exemplar-free CIL approach that leverages confidence scores obtained from the probability distribution (logits) of the classifier. The proposed method is based on the idea that the information stored in the confidence, expressed in probability distribution of a classifier, can be used to determine whether a given data point belongs to a specific task when compared to the probability distribution provided by another classifier. To ensure that expert classifiers produce high-entropy distributions for data unrelated to their specific task, we employ a unique loss function known as DisMax loss \cite{DBLP:journals/corr/abs-2205-05874} during the training phase. CTP is designed to determine the task-id by defining a noise region in the distribution and computing a confidence score based on the number of logits falling within this region. By analyzing logits distribution, CTP effectively discerns between in-distribution and out-of-distribution data, enabling accurate task-id prediction.


In our experimental evaluation, we concentrate on two key disease classification datasets: PathMNIST \cite{kather2019predicting} and Optical Coherence Tomography (OCT) \cite{kermany2018large}. We demonstrate that CTP outperforms other relevant methods, indicating its effectiveness in handling the challenge of unknown task identities at inference time. Medical imaging procedures often involve the acquisition of multiple images to capture different aspects of the anatomy, pathology, or physiological processes. Capitalizing on this, the accuracy of CTP improves if a continuum of image is available at the time of inference.
By introducing CTP, we contribute to the development of robust and adaptive continual learning systems that can handle unknown task identities at inference time for continual disease classification.

\section{Related Work}
CL methods are broadly categorized into three types. \textit{Regularization-based} methods \cite{zenke2017continual,dhar2019learning,goswami2023fecam} add a penalty to the loss function to preserve knowledge from past tasks while learning new ones. \textit{Replay-based} techniques \cite{rebuffi2017icarl,buzzega2020dark,van2020brain} replay a subset of previous data to mitigate forgetting. \textit{Expansion-based} strategies \cite{rusu2016progressive,xu2018reinforced}, relevant to our work, increase the model's capacity for new tasks by adding parameters or creating task-specific models. PNN \cite{rusu2016progressive} introduce new columns for each task, DEN \cite{yoon2018lifelong} dynamically retrain and expand the network, RCL \cite{xu2018reinforced} applies reinforcement learning for expansion, and APD \cite{yoon2020apd} combines shared and task-specific parameters. These methods, which rely on known task identities at inference, have limited use in disease classification due to this assumption.

Several strategies predict task identity to leverage task-specific models or parameters for classification. EFT \cite{verma2021efficient} generates task-specific features through specialized convolutions and regularization for improved task prediction. $H^2$ \cite{jin2022helpful} uses a meta-classifier and contrastive loss to distinguish between tasks. iTAML \cite{rajasegaran2020itaml} is an exemplar-based meta-learning method that computes scores for each task to predict the task identifier, assuming a data continuum scenario where all data samples at inference time belong to the same task. SupSup \cite{wortsman2020supermasks} learns task-specific masks by formulating the task identification problem as a gradient-based optimization problem. CLOM \cite{kim2022continual}, fitting both task and class incremental learning, learns task-specific masks and employs out-of-distribution detection for classifier training. L2P \cite{wang2022learning} learns to dynamically prompt (task-specific learnable parameters) a pre-trained model to learn tasks sequentially. In contrast to these methods, our method CTP determines the task identifier by comparing the output logits of the task-specific classifiers. CTP functions based on the principle that employing an expert network for each task leads to improved accuracy. Our experimental results on disease classification datasets demonstrate that CTP outperforms these existing methods by a significant margin. 

\section{Confidence-based Task Prediction (CTP)}

 Continual learning involves training machine learning models on data from a series of tasks $\mc{D}=\{\mc{D}_1, \mc{D}_2,\cdots,\mc{D}_T\}$. Task $\mc{D}_t=\{(\boldmath{x}^t_i, y^t_i)\}_{i=1}^{n_t}$ is the $t^{th}$ task where $\boldmath{x}^t_i\in X_t$ is an input, $y^t_i\in Y_t$ is the corresponding label and $n_t$ is number of classes in the task. During training the $t^{th}$ task, only training data $\mc{D}_t$ is available while the training data of previous tasks are no longer accessible. The tasks are generally assumed to be distinct from one another. The goal of continual learning is to train a parameterized model $f : X \rightarrow Y$ that can predict the correct label for an unseen test sample from any task.
Our proposed approach, CTP, is based on the concept that the determination of whether a data point belongs to a specific expert classifier can be achieved by comparing the probability distribution outputs (logits) of the classifiers. The concept is that for a given input belonging to task $t$, the classifier $f_t$ produces higher logits for the correct class and lower logits for incorrect classes, creating a shortage of logits with intermediate values. 
\begin{figure}
    \centering
    \subfloat[\label{fig:dist}] {\includegraphics[scale=0.4]{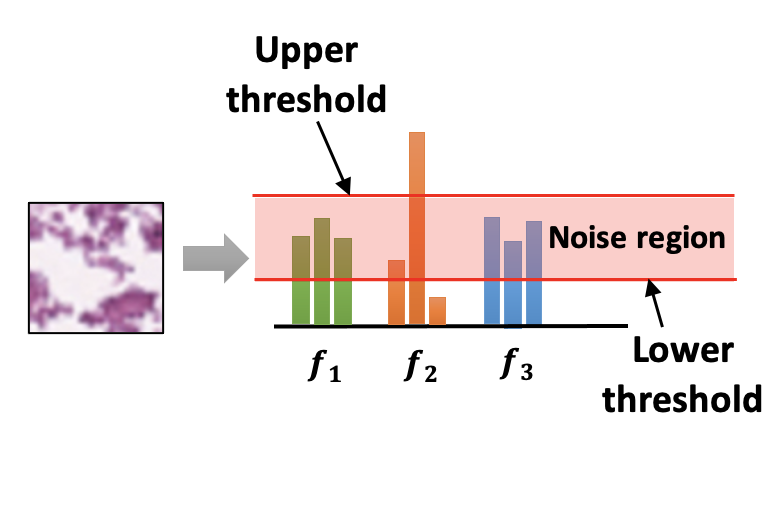}}
    \subfloat[\label{fig:noise}] {\includegraphics[scale=0.28]{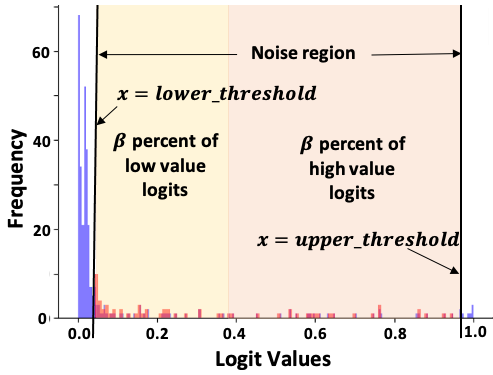}}
    \caption{\protect\subref{fig:dist} The correct classifier ($f_2$) for the PathMNIST image produces high-value logits for the correct class and low-value logits for the incorrect classes. The logits from all the classifier outputs are combined, and a noise region (in red) is established. The confidence score for each classifier is computed based on the number of logits from that classifier falling within the defined noise region. \protect\subref{fig:noise} Noise region is determined based on histogram of the combined logits. $\beta$ percent of logits are chosen from both high-value and low-value logits to define the noise region.}
\end{figure}
Figure \ref{fig:dist} illustrates an example where the correct classifier ($f_2$) for the PathMNIST image exhibits a high logit value for the correct class and low logit values for incorrect classes. Conversely, the logit for incorrect classifiers have intermediate values. We term this intermediate range as the noise region and calculate the confidence score based on the number of logits falling within this region for each classifier. The classifier with the least number of logits in the noise region is identified as the correct classifier for the input data. For each task with task-id $t$, an expert classifier $f_t(\boldmath{x})$ is trained. We define $C_t(\boldmath{x})$ as a score that represents confidence that data point $\boldmath{x}$ belongs to classifier $f_t$. The problem of classification of a data from unknown task can be solved by first determining the task-id $k=\operatorname*{argmax}_{t} C_t(\boldmath{x})$ and then using the expert classifier $f_k$ to determine $\hat{y}=\operatorname*{argmax}_{n_k} f_k(\boldmath{x})$, the class of the input data. 
\subsubsection{Learning the Classifier:}
The minimization of cross-entropy loss is a widely used approach for training a classifier. However, cross-entropy loss has a tendency to generate high confidence predictions, even for incorrect data that does not align with the distribution of the training data \cite{guo2017calibration}. To enable classifiers to effectively differentiate between in-distribution data and out-of-distribution data (data belonging to the different task classifier), we employ DisMax loss function \cite{DBLP:journals/corr/abs-2205-05874} during the classifier training process. It relies on the notion that a classifier should generate a probability distribution with high entropy for data that does not belong to its assigned task. DisMax incorporates mean isometric distance \cite{macedo2021enhanced} across all prototypes in addition to the specific isometric distance for each class, to get the updated logits ($L^j$). The loss is computed as follows
\begin{align}
\mc{L}_{DisMax} = - log\Big(\frac{exp(E_s L^k)}{\sum_j exp(E_s L^j)}\Big), L^j = -\Big(D^j+\dfrac{1}{N} \sum_{n=1}^{N}D^n\Big)
\end{align}

Here, $E_s$ represents an entropic scale \cite{macedo2021enhanced} and $L^j$ is the enhanced logits of class $j$. $D^j=|d_s|\ ||\widehat{f_{\theta}(x)}-\widehat{p^j_{\phi}}||$ is the isometric distance relative to class $j$. $|d_s|$ denotes the absolute value of the learnable scalar termed distance scale \cite{macedo2021enhanced}, $f_{\theta}(x)$ is the original logits and $p^j_{\phi}$ represent the learnable prototype associated with class $j$. Expression $||v||$ represents 2-norm of vector $v$ and $\hat{v}$ represents 2-norm normalization of $v$.
\subsubsection{Noise region:}
Figure \ref{fig:dist} presents a high-level framework for identifying the noise region in our approach. The goal is to accurately determine the region of logits that represents uncertainty or noise in the classification process. To achieve this, we start by aggregating all the logits produced by each classifier for a given input. The distribution of logits for the correct classifier is expected to be skewed because for each high-value logit (indicating the correct class), there will be $n_t-1$ low-value logits (indicating incorrect classes) when there are $n_t$ classes for task $t$. Rather than opting for a straightforward middle logit range selection (e.g., between the 5th and 95th percentiles), we introduce two hyperparameters, namely $\alpha$ and $\beta$, to determine the noise region. The parameter $\alpha$ represents the percentile threshold above which we consider the logits to have high values, indicating that the classifier is confident about the predicted class. This separation divides the logits into two groups: high values and low values.

To define the noise region, we aim to exclude highly confident logits, indicative of either very high values for the correct class or very low values for incorrect classes. Therefore, we use the parameter $\beta$, representing the percentage of logits to consider from both the high-value and low-value regions. By appropriately setting the values of $\alpha$ and $\beta$, we identify the noise region in the logits distribution. We determine \textit{upper-percentile} $= \alpha + (1-\alpha) \beta$ and \textit{lower-percentile} $= \alpha - \alpha \beta$. We then calculate the corresponding percentile values (\textit{upper-threshold} and \textit{lower-threshold}) to establish the range of logit values within which the logits are categorized as noise. The histogram displayed in Figure \ref{fig:noise} represents the combined logits, where high-confidence logits are depicted in blue and noisy logits are indicated in red. The left shaded region encompasses $\beta$ percent of logits from the low-value region, while the right shaded region comprises $\beta$ percent of logits from the high-value region.  

\subsubsection{Confidence computation:}

\begin{algorithm} [h]
\caption{Confidence Computation}
\label{algo-cs}
\hspace*{\algorithmicindent} \textbf{Input:} logits $f(\boldmath{x})$, \textit{upper-threshold}, \textit{lower-threshold} \\
 \hspace*{\algorithmicindent} \textbf{Output:} confidence score $C(\boldmath{x})$
\begin{algorithmic}[1]
\STATE{num-class = len($f(\boldmath{x})$)}
\STATE{$C(\boldmath{x})=0$}
 \FOR{ $v$ in $f(\boldmath{x})$}
 \IF{$\textit{lower-threshold} \le v \le \textit{upper-threshold}$}
 \STATE{$C(\boldmath{x}) = C(\boldmath{x}) + 1$}
 \ENDIF
 \ENDFOR
 \STATE {$C(\boldmath{x}) = \frac{C(\boldmath{x})}{\text{num-class}}$
 \RETURN $C(\boldmath{x})$}
\end{algorithmic}
\end{algorithm} 
After determining the noise region, the confidence for each classifier is computed based on number of logits falling in the noise region. Since the confidence score in CTP is determined by the quantity of logits found within the noise region, having multiple images from the same task during inference (data continuum) proves advantageous as it increases the overall count of logits available for analysis.  
In order to ensure comparable logit values, we normalize the logit values of each distribution independently. 
We utilize the threshold values on the normalized logits of each classifier to calculate the confidence score. The individual confidence score is normalized by dividing it by the number of classes. Algorithm \ref{algo-cs} outlines the steps for computing the confidence score.  

\section{Experiment and Results}
\begin{figure}[b]
    \centering
    \includegraphics[scale=0.22]{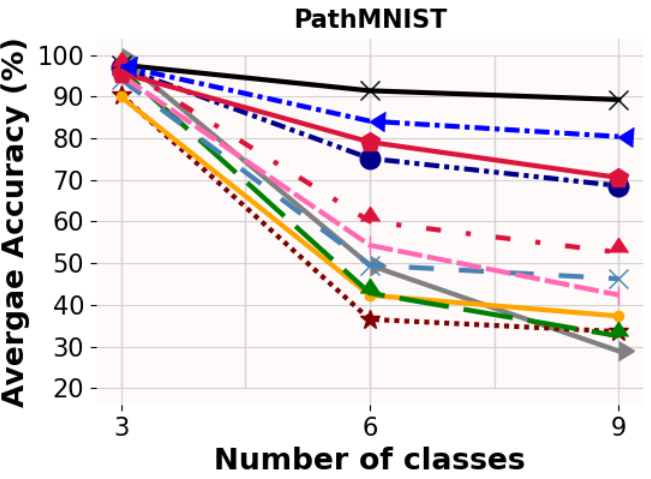}
    \includegraphics[scale=0.22]{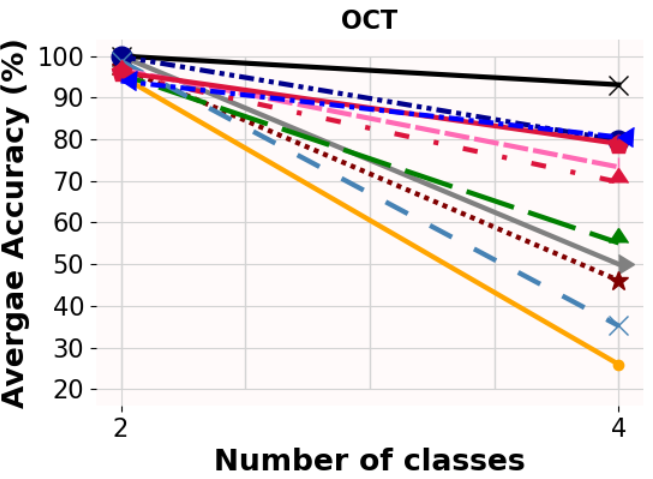}
    \includegraphics[scale=0.22]{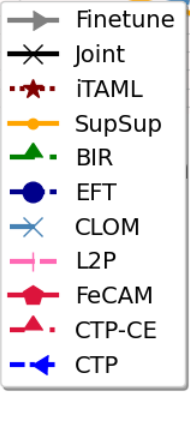}
    \caption{Average classification accuracy as training progresses.}
\label{fig:overall}
\end{figure}
In our experimental evaluation, we focus on two prominent disease classification datasets: PathMNIST \cite{kather2019predicting} and OCT \cite{kermany2018large}. PathMNIST is composed of histology slides categorized into nine distinct classes of colon pathologies, while the OCT encompasses four types of retinal diseases spread across four classes. For the purposes of our study, the nine classes in PathMNIST were segmented into three tasks, with each task including three classes. Similarly, the OCT dataset was organized into two tasks, each consisting of two classes.
We compare CTP with several state-of-the-art continual learning methods, including iTAML \cite{rajasegaran2020itaml}, SupSup \cite{wortsman2020supermasks}, BIR \cite{van2020brain}, EFT \cite{verma2021efficient}, CLOM \cite{kim2022continual}, L2P \cite{wang2022learning}, and FeCAM \cite{goswami2023fecam}. Similar to CTP, iTAML, SupSup, EFT, CLOM, and L2P initially predict the task-id prior to executing the final classification. CTP-CE is a variant of CTP where instead of DisMax loss function, cross-entropy loss is used to train the classifiers. The finetune approach, the lower baseline, optimizes weights for new tasks without considering performance on past tasks. Joint training, the upper baseline, trains the model with current and all previous tasks' data combined.

\begin{table}[t]
\begin{center}
\caption{Classification accuracy of each task on the final model. The data continuum size for iTAML, CTP-CE and CTP is set to 5.}
\begin{tabular}{|l|cccc|ccc|}
\hline
&\multicolumn{4}{c|}{PathMNIST} & \multicolumn{3}{c|}{OCT}\\
Method & Task 1 & Task 2 & Task 3 & Average & Task 1 & Task 2 & Average\\
\hline
Finetune & 0.00 & 0.00 & 86.70 & 28.89 & 0.00 & 100.0 & 50.00 \\
Joint & 95.99 & 85.93 & 85.92 & 89.28 & 99.47 & 86.73 & 93.10 \\ \hline
iTAML & 32.77 & 30.34 & 37.59 & 33.57 & 46.07 & 46.13 & 46.10\\
SupSup & 32.37 & 40.74 & 38.78 & 37.29 & 26.00 & 25.87 & 25.93 \\
BIR  & 0.00 & 1.96 & \textbf{95.52} & 32.49 & 18.80 & \textbf{91.40} & 55.10 \\ 
EFT  & 62.23 & 66.82 & 76.72 & 68.59 & 76.63 & 82.87 & 79.75 \\
CLOM  & 44.29 & 49.60 & 44.90 & 46.26 & 35.33 & 35.07 & 35.20 \\ 
L2P  & 23.04 & 31.04 & 72.97 & 42.35 & 69.86 & 76.79 & 73.33 \\
FeCAM  & \textbf{82.50} & 68.30 & 60.79 & 70.53 & 77.13 & 80.73 & 78.93 \\
CTP-CE  & 63.92 & 66.82 & 76.72 & 68.59 & 70.53 & 68.87 & 69.70 \\
CTP  & 77.97 & \textbf{75.05} & 88.18 & \textbf{80.40} & \textbf{85.40} & 75.60 & \textbf{80.50} \\
\hline
\end{tabular}
\label{tab-acc}
\end{center}
\end{table}

We employed the ResNet50 \cite{he2016deep} architecture as the base model for our classifiers and trained them for 64 epochs using a learning rate of 1e-3. The data continuum size was set to 5, while $\alpha$ and $\beta$ were assigned values of 0.25 and 0.75, respectively. The reported results are the average of three runs with different random seeds.

Figure \ref{fig:overall} depicts the drop in average accuracy for both the datasets as the learning progresses for subsequent tasks. While the initial accuracy of most methods is similar, as the learning progresses, many of them experience decline in average accuracy due to degraded performance on older tasks. Table \ref{tab-acc} showcases the classification accuracies achieved for each task upon completing the learning of all tasks. CTP exhibits consistent performance across both datasets, 
\begin{figure} [h]
    \centering
    \includegraphics[scale=0.28]{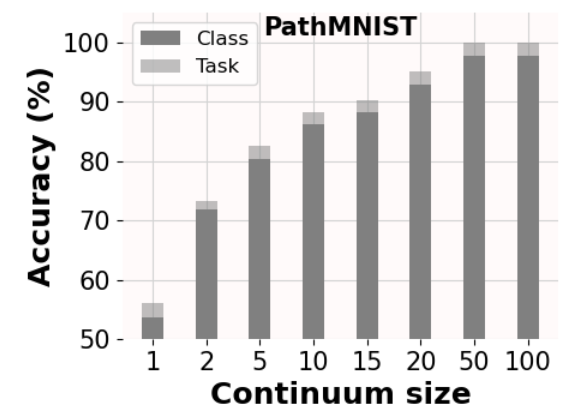}
    \includegraphics[scale=0.28]{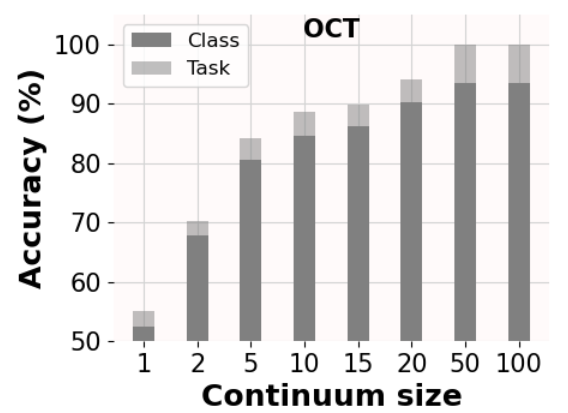}
    \caption{Final average task prediction accuracy and classification accuracy for different sizes of data continuum.}
\label{fig:cont}
\end{figure}
achieving an accuracy of 80.4\% on PathMNIST and 80.5\% on OCT. Notably, CTP maintains a balanced accuracy distribution across all tasks, distinguishing itself from certain methods like BIR and L2P, which display imbalances with higher accuracy for newer tasks and lower accuracy for older tasks. iTAML, also dependent on data continuum for task-id prediction, exhibits subpar performance on disease classification datasets with a continuum size of 5. Although FeCAM and EFT demonstrate good performance on OCT data, their efficacy diminishes when applied to PathMNIST. This suggests that these methods may not be well-suited for the task of continual disease classification.

\begin{figure} [h]
    \centering
    \includegraphics[scale=0.38]{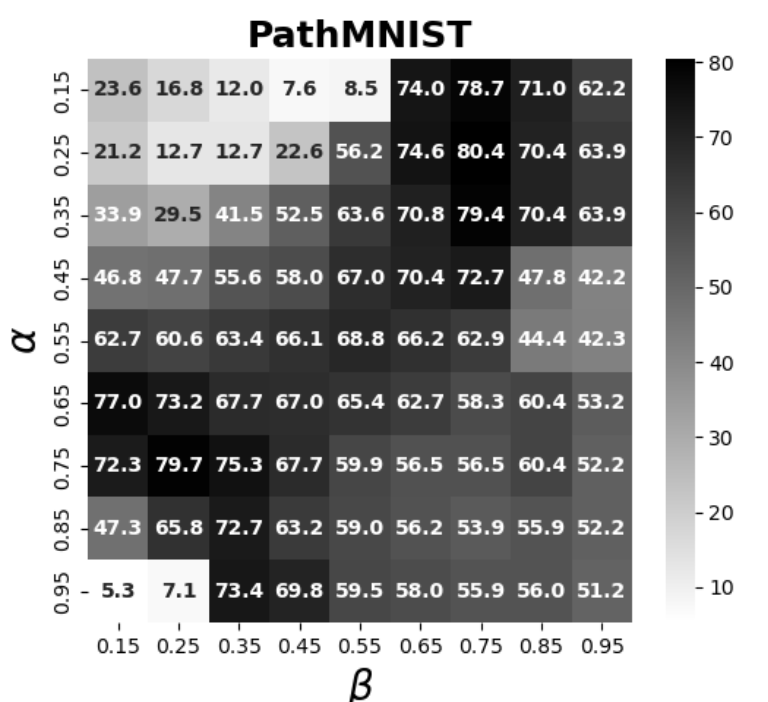}
    \includegraphics[scale=0.38]{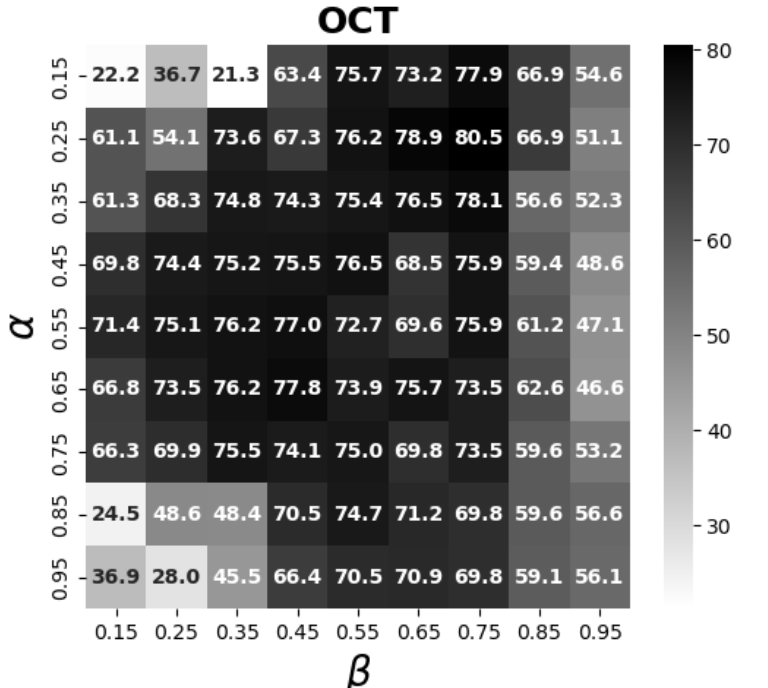}
    \caption{Classification accuracies across various combinations of $\alpha$ and $\beta$ values.}
\label{fig:ablation}
\end{figure}

Figure \ref{fig:cont} illustrates the impact of data continuum size on CTP's accuracy in both task prediction and classification.
A classification accuracy of approximately 70\% is attained using only two images, underscoring the applicability of CTP in disease classification scenarios where multiple images are typically acquired to capture diverse aspects of anatomy. Remarkably, with data continuum sizes of 15 and above, the accuracy closely matches that of joint training. This suggests that in scenarios where a data continuum is present during inference, CTP performs equivalently to joint training. Figure \ref{fig:ablation} illustrates the impact of different $\alpha$ and $\beta$ values, revealing that values within the range of 0.25-0.35 for $\alpha$ and 0.65-0.75 for $\beta$ lead to improved results across both datasets. As a result, the noise region typically corresponds to logits within the range of approximately the $10^{th}$ to $80^{th}$ percentiles.

\section{Conclusion}
We introduced, CTP, a confidence-based task-id predictor, tackling the issue of unrecognized task identities in continual learning. CTP, an exemplar-free approach to class incremental learning (CIL), is based on the premise that having an expert network for each task is preferable if the task-id can be identified at the time of inference. CTP leverages classifier logits to define a noise region for precise task-id identification, showcasing superior accuracy compared to existing methods. The modification of classifier logits through a specialize loss function promotes high-entropy distributions for data associated with other tasks. Additionally, CTP exhibits enhanced performance when provided with data continuums during inference, achieving accuracy comparable to joint training in such scenarios. Our findings highlight the crucial role of task-id prediction in continual disease classification and establish CTP's superiority in managing unknown task identities. 
 

%
%
%
\bibliographystyle{splncs04}
\bibliography{ctp}

\end{document}


\title{Confidence-Based Task Prediction in Continual Disease Classification Using Classifier Probability Distribution}
%
%
%
\authorrunning{Paper 305}
%
%
%

\section{Introduction}
The steps to compute confidence score has been outlined in Algorithm \ref{algo-cs}
\begin{algorithm} [h]
\caption{Confidence Computation}
\label{algo-cs}
\hspace*{\algorithmicindent} \textbf{Input:} logits $f(\boldmath{x})$, \textit{upper-threshold}, \textit{lower-threshold} \\
 \hspace*{\algorithmicindent} \textbf{Output:} confidence score $C(\boldmath{x})$
\begin{algorithmic}[1]
\STATE{num-class = len($f(\boldmath{x})$)}
\STATE{$C(\boldmath{x})=0$}
 \FOR{ $v$ in $f(\boldmath{x})$}
 \IF{$\textit{lower-threshold} \le v \le \textit{upper-threshold}$}
 \STATE{$C(\boldmath{x}) = C(\boldmath{x}) + 1$}
 \ENDIF
 \ENDFOR
 \STATE {$C(\boldmath{x}) = \frac{C(\boldmath{x})}{\text{num-class}}$
 \RETURN $C(\boldmath{x})$}
\end{algorithmic}
\end{algorithm} 

\begin{figure}
    \centering
    \includegraphics[scale=0.35]{bar-alpha.png}
    \caption{Final average task prediction accuracy and classification accuracy for different values of $\alpha$. Data continuum was set to 5 and $\beta$ value was fixed to 0.85.}
    \label{fig:alpha}
\end{figure}
\begin{figure}
    \centering
    \includegraphics[scale=0.35]{bar-beta.png}
    \caption{Final average task prediction accuracy and classification accuracy for different values of $\beta$. Data continuum was set to 5 and $\alpha$ value was fixed to 0.75.}
    \label{fig:beta}
\end{figure}
\begin{figure}
    \centering
    \includegraphics[scale=0.3]{bar-continuum.png}
    \caption{Final average task prediction accuracy and classification accuracy for different sizes of data continuum. $\alpha$ and $\beta$ values were fixed at 0.75 and 0.85 respectively.}
    \label{fig:continuum}
\end{figure}
\subsubsection{Ablation Studies:}
In this section, we conduct an ablation study to investigate the impact of different values of $\alpha$, $\beta$, and data continuum size. Figure \ref{fig:alpha} presents the accuracies obtained when $alpha$ values were set to 0.85, 0.75, 0.65, and 0.55, while keeping $\beta$ fixed at 0.85. Our experiments indicate that the best result is achieved when $\alpha$ is set to 0.75, meaning that logits below the 75th percentile are considered as low-value logits, while logits above the 75th percentile are considered as high-value logits. 

Figure \ref{fig:beta} presents the results of the ablation study on different $\beta$ values. We explored the values of 0.35, 0.55, 0.75, and 0.95 while keeping the $\alpha$ value fixed at 0.75. Our observations indicate that the best result is obtained around the value of 0.85, and the performance starts to decline if we increase it further. Therefore, to define the noise region, we selected 85\% of the logits each from the high-value logit group and the low-value logit group.

In Figure \ref{fig:continuum}, we present the accuracies for different values of the data continuum size. For CTP, we achieved a task prediction accuracy of 63.36\% and a classification accuracy of 61.92\% for data continuum size of 1. Therefore, CTP achieves comparable accuracy to EFT even when the data continuum size is set to 1. When the data continuum size was set to 2, the task prediction accuracy was 73.15\% and the classification accuracy was 70.46\%. Remarkably, for data continuum sizes of 15 and above, the task prediction accuracy reached 100\%, indicating that in a scenario where data continuum is available during inference, CTP performs as well as joint training. It's worth noting that iTAML requires a data continuum size of 100 (as reported in the iTAML paper) to achieve 100\% task prediction accuracy for their setting.

During our training, one of the valued we encountered for thresholds when $\alpha = 0.25$ and $\beta=0.75$ are as follows.
\begin{align*}
    &up = 0.25 + (1-0.25) * 0.75 = 0.8125, &ut = 81.25th \text{ percentile} = 0.799 \\
    &lp = 0.25 - 0.25 * 0.75 = 0.0625, &lt = 6.25th \text{ percentile} = 0.018
\end{align*}

Hence all the logits whose values fell between 0.0017 and 0.643 were considered as noise for that experiment.
\bibliographystyle{splncs04}
\bibliography{ctp}